\documentclass{article}
\usepackage{spconf}
\usepackage{newtxtext,newtxmath}
\usepackage{amsmath}
\usepackage{graphicx}
\usepackage{booktabs,multirow}
\usepackage{cite}
\usepackage{microtype}
\usepackage{xspace}
\usepackage[hyphens]{url}
\usepackage[table]{xcolor}
\usepackage{array}
\newcommand{\method}{Diffusion-Grounded VideoLLM\xspace}
\newcommand{\dtl}{DTL\xspace}

\title{DIFFUSION-GROUNDED VIDEOLLM FOR ENTITY-AWARE TEMPORAL GROUNDING}

\name{Pengcheng Fang$^{2}$\sthanks{These authors contributed equally.}, Yuxia Chen$^{3,*}$, Yong Wang$^{3}$, Xiaohao Cai$^{2}$, Xiangning Ruan$^{1}$\sthanks{Corresponding author.}}
\address{
$^{1}$Department of Electronic Engineering , Tsinghua University, China\\
$^{2}$University of Southampton, Southampton, UK\\
$^{3}$University of Electronic Science and Technology of China, Chengdu, China
}

\begin{document}
\ninept
\maketitle

\begin{abstract}
Precise temporal grounding requires distinguishing when a queried event occurs from when its participating entities are merely visible. We propose \method, which conditions temporal feature extraction on query-relevant entities before language reasoning. The framework tracks entities named in the query and uses their masks to condition a frozen video diffusion backbone. Intermediate spatiotemporal features are extracted through truncated denoising and combined with entity tokens and timestamp embeddings. The language model uses this evidence together with the full query to generate temporal intervals and answers to grounded questions. On Charades-STA and NExT-GQA, the model obtains 43.5 mIoU and 28.4 Acc@GQA, improving the reported Grounded-VideoLLM reference by 6.7 and 1.7 points, respectively. Component and entity-pathway ablations support the usefulness of conditioning diffusion features on query-relevant entities for temporal grounding.
\end{abstract}


\section{Introduction}
\label{sec:introduction}
Video temporal grounding identifies the interval corresponding to a language-described event in an untrimmed video~\cite{gao2017tall}. Precise localization requires distinguishing the occurrence of an event from the presence of its participants. For example, in a query such as ``When does the cat climb onto the man?'', both the cat and the man may be visible before and after the climbing action. Recognizing these entities therefore provides only part of the evidence: the model must associate them with the motion and interactions that define the queried event. A relevant boundary may correspond to a change in posture or contact, even though the identities of the participants remain unchanged. This motivates a temporal representation that connects query-relevant entities to their visual changes over time.

Video large language models (VideoLLMs) support language-based video understanding by mapping visual evidence into a language-model interface. Existing temporal grounding methods make time more accessible through normalized frame indices, timestamp-aware features, or discrete time tokens~\cite{huang2024vtimellm,ren2024timechat,wang2024grounded}. These mechanisms establish a correspondence between video evidence and temporal positions, but explicit timestamps alone do not determine which entities the evidence should describe. In particular, when visual encoding is independent of the query, entity selection must be resolved from features already extracted for the whole scene. The representation must support both selecting the relevant participants and identifying the transitions that distinguish the described action from its surrounding context. Conditioning feature extraction on the queried entities offers a complementary way to organize the evidence used for temporal localization.

Pixel-level grounding provides spatial evidence for this condition. Grounding DINO localizes text-referred regions~\cite{liu2024groundingdino}, enabling query-relevant entity tracks to guide temporal feature extraction. Tracking carries these spatial references across frames, helping associate local observations with the same query referent as its appearance and position change. Video diffusion backbones learn appearance and temporal structure through conditional denoising, while editing frameworks such as VACE accept spatial masks as conditions~\cite{wan2025wan,jiang2025vace}. These capabilities motivate conditioning a pretrained diffusion backbone on entity masks and extracting intermediate features for temporal grounding without decoding a generated video.

We propose \method, which conditions video feature extraction on query-relevant entity tracks before language reasoning. Entity masks condition a frozen video diffusion backbone, and intermediate features from a truncated denoising process form a diffusion temporal latent (\dtl). Spatial pooling retains the temporal dimension, allowing these features to be combined with timestamp embeddings; separate entity tokens provide additional information about the tracked participants. A language model adapted with low-rank updates uses this representation to generate interval timestamps and answers to grounded questions. Throughout this process, entity tracks provide auxiliary visual evidence, while the complete query retains the action and temporal semantics needed to determine which interval to generate.

Our contributions are threefold: (1) an entity-conditioned temporal grounding framework that introduces query-relevant tracks into video feature extraction; (2) a diffusion temporal representation integrated with entity tokens and timestamp embeddings for interval generation and grounded question answering; and (3) evaluation on Charades-STA, DiDeMo, and NExT-GQA, with component and entity-pathway ablations and analyses of extraction position, tracking, and representation budget.

\section{Method}
\label{sec:method}
\begin{figure*}[t]
\centering
\includegraphics[width=\textwidth]{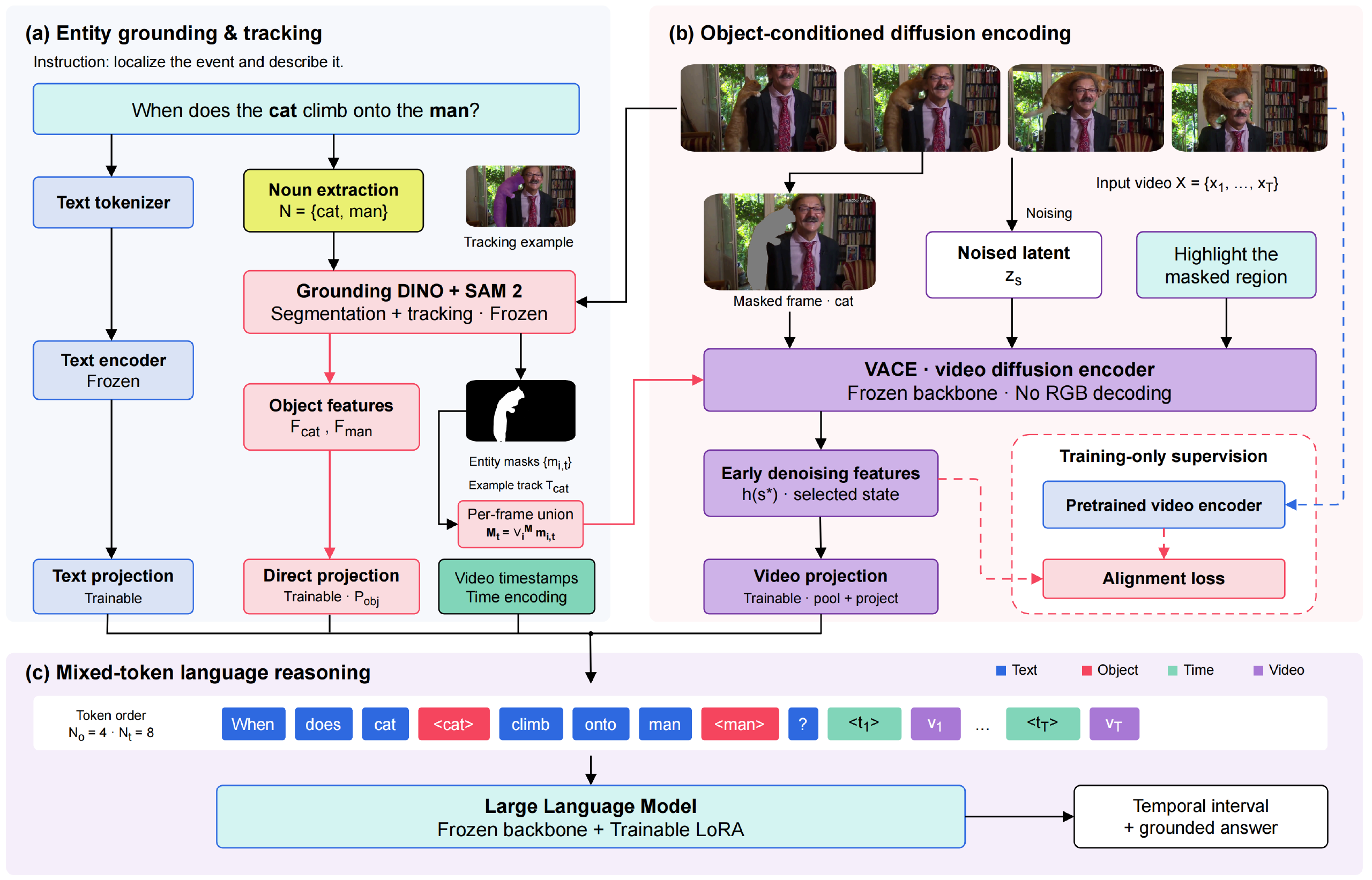}
\caption{Overview of \method. Query-derived tracks supply entity tokens and masks that condition VACE. Early diffusion features receive alignment supervision during training. Visual, text, and diffusion features are concatenated with timestamp information for interval and answer generation. Token order is schematic.}
\label{fig:framework}
\end{figure*}

Given a video $\mathcal V=\{(x_t,s_t)\}_{t=1}^{T}$ sampled at 1\,fps, with source timestamps $s_t$ in seconds, and a query $q$, \method generates the event's start and end times and, for grounded VideoQA, a textual answer. Figure~\ref{fig:framework} shows how entity evidence conditions temporal feature extraction before language reasoning.

\subsection{Query-conditioned entity evidence}
\label{subsec:tracking}
We extract visual noun phrases $\{n_i\}_{i=1}^{M}$ for detection and retain the full query, including action and temporal expressions, for reasoning. Grounding DINO~\cite{liu2024groundingdino} supplies region proposals and text--region matching scores. Let $v_{i,t}$ indicate that the highest-scoring proposal for referent $i$ passes its detection threshold at frame $t$, with $v_{i,t}=0$ when no proposal is available. Tracking starts at the earliest persistent co-occurrence:
\begin{equation}
t_s=\min\left\{t:1\leq t\leq T-K+1,\;
\prod_{u=0}^{K-1}\prod_{i=1}^{M}v_{i,t+u}=1\right\},
\end{equation}
when the set is nonempty and $M\geq1$. Grounded-SAM initializes region masks, and SAM~2 propagates them~\cite{ren2024grounded,ravi2024sam2}.

Let $\mathcal A_t$ index tracks with available masks at frame $t$. Their union, $M_t=\bigvee_{i\in\mathcal A_t}m_{i,t}$, conditions diffusion encoding. Separate track features provide entity tokens:
\begin{equation}
o_i=g_o\!\left(\operatorname{Pool}_{t,h,w}(F_t\odot m_{i,t})\right),
\end{equation}
where $F_t$ is the grounding feature grid, masks are aligned to that grid, and pooling aggregates available track observations. Missing observations provide no auxiliary entity evidence; the video and full query remain available. Co-occurrence initializes tracking rather than determining event boundaries: the full query specifies the action to localize.

\subsection{Diffusion temporal features}
\label{subsec:dtl}
We condition VACE~\cite{jiang2025vace,wan2025wan} on the video, available masks, and a short instruction identifying the regions of interest. Features are extracted within the early 20\% of reverse denoising, before full video synthesis; no RGB video is decoded. This fraction denotes progress along the denoising trajectory rather than a fixed number of evaluations.

Let $H^{(\ell)}_{\rho}$ be the activation grid at layer $\ell$ and denoising progress $\rho$. At the selected early state $\rho_*$, we average later transformer layers:
\begin{equation}
\bar H_{\rho_*}=\frac{1}{|\mathcal L_{\rm late}|}
\sum_{\ell\in\mathcal L_{\rm late}}H^{(\ell)}_{\rho_*},\qquad
z_j=g_d\!\left(\operatorname{Pool}_{h,w}(\bar H_{\rho_*,j})\right).
\end{equation}
The sequence $\{z_j\}_{j=1}^{T'}$ forms the diffusion temporal latent (\dtl), with $g_d$ projecting features to the language-model dimension. Spatial pooling retains the native temporal axis, whose length $T'$ may differ from the sampled-frame count because of temporal compression. Each slice is associated with its source-video timestamp $\tilde s_j$. Layer averaging aggregates network depth while preserving video time.

\subsection{Mixed features and learning}
\label{subsec:mixedtokens}
For compact input, temporal features are pooled into $N_t$ bins and entity descriptors are compressed into $N_o$ slots, where $N_t$ and $N_o$ denote the numbers of temporal bins and object slots, respectively. These budgets control the input representation. The mixed interface concatenates projected visual features, text features, and VACE-derived diffusion features along the sequence dimension. Entity descriptors supply visual tokens, while timestamp embeddings $e_j^{\rm time}=\operatorname{Enc}_{\rm time}(\tilde s_j)$ associate temporal features with source-video time. Denoting the resulting feature groups by $Z_{\rm vis}$, $Z_{\rm text}$, and $Z_{\rm diff}$, the language-model context is
\begin{equation}
C=\operatorname{Concat}_{\rm seq}(Z_{\rm vis},Z_{\rm text},Z_{\rm diff}).
\end{equation}
The full query is retained in $Z_{\rm text}$. The model generates a target response $y$ containing interval timestamps and an answer when required, following autoregressive VideoLLM grounding~\cite{huang2024vtimellm,ren2024timechat,wang2024grounded}:
\begin{equation}
\mathcal L_{\rm gen}=-\sum_{k=1}^{|y|}\log p_\psi(y_k\mid y_{<k},C).
\end{equation}
The predicted interval is read directly from the generated response.

Following REPA~\cite{yu2024repa}, representation alignment supplies a visual prior to the extracted early diffusion features using clean-input representations from a pretrained visual encoder. Denote this feature-alignment objective by $\mathcal L_{\rm align}$. Training combines response and representation supervision:
\begin{equation}
\mathcal L=\mathcal L_{\rm gen}+\lambda_{\rm align}\mathcal L_{\rm align}.
\end{equation}
The grounding model, VACE backbone, and base language-model weights remain frozen; feature projections and LoRA~\cite{hu2022lora} parameters are optimized. The reference encoder supplies training-only supervision.

\section{Experiments}
\label{sec:experiments}

\begin{table}[t]
\centering
\footnotesize
\setlength{\tabcolsep}{1.3pt}
\renewcommand{\arraystretch}{1.03}
\caption{\textbf{Temporal grounding on Charades-STA and DiDeMo.}
R@1 is reported at different IoU thresholds.}
\label{tab:tsg_results}
\begin{tabular*}{\columnwidth}
{@{\extracolsep{\fill}}l|c|cccc|cccc@{}}
\toprule
\multirow{2}{*}{Model}
& \multirow{2}{*}{Size}
& \multicolumn{4}{c|}{Charades-STA}
& \multicolumn{4}{c}{DiDeMo} \\
\cmidrule(lr){3-6}\cmidrule(lr){7-10}
& & R.3 & R.5 & R.7 & mIoU
& R.3 & R.5 & R.7 & mIoU \\
\midrule
Video-LLaMA~\cite{zhang2023videollama}
& 7B & 25.2 & 10.6 & 3.4 & 16.8
& 20.1 & 8.2 & 2.5 & 14.3 \\

SeViLA~\cite{yu2023sevila}
& 3B & 27.0 & 10.5 & 5.8 & 18.3
& 23.5 & 9.8 & 3.6 & 15.9 \\

Video-ChatGPT~\cite{maaz2024videochatgpt}
& 7B & 27.2 & 6.2 & 1.9 & 19.7
& 19.8 & 6.5 & 1.2 & 13.7 \\

Momentor~\cite{qian2024momentor}
& 7B & 42.6 & 26.6 & 11.6 & 28.5
& 38.2 & 21.8 & 9.4 & 26.5 \\

VTimeLLM~\cite{huang2024vtimellm}
& 7B & 51.0 & 27.5 & 11.4 & 31.2
& 45.0 & 28.8 & 12.0 & 27.9 \\

TimeChat~\cite{ren2024timechat}
& 7B & 47.7 & 22.9 & 12.5 & 31.6
& 42.8 & 24.4 & 11.3 & 28.2 \\

HawkEye~\cite{wang2024hawkeye}
& 7B & 50.6 & 31.4 & 14.5 & 33.7
& 44.8 & 29.7 & 13.4 & 29.5 \\

Grounded-VideoLLM~\cite{wang2024grounded}
& 3.8B & 54.2 & 36.4 & 19.7 & 36.8
& 48.6 & 33.4 & 15.9 & 32.0 \\

LLaVA-ST~\cite{li2025llavast}
& 7B & \textbf{63.1} & \underline{44.8}
& \textbf{23.4} & \underline{42.4}
& \textbf{56.2} & \underline{39.8}
& \underline{20.1} & \underline{37.6} \\

\rowcolor{gray!15}
Ours
& 3.8B & \underline{58.7} & \textbf{45.2}
& \underline{23.0} & \textbf{43.5}
& \textbf{56.2} & \textbf{40.0}
& \textbf{20.5} & \textbf{38.2} \\
\bottomrule
\end{tabular*}
\end{table}

\begin{table}[t]
\centering
\footnotesize
\setlength{\tabcolsep}{2.6pt}
\renewcommand{\arraystretch}{1.03}
\caption{\textbf{Grounded VideoQA results on NExT-GQA.}}
\label{tab:grounded_videoqa}
\begin{tabular*}{\columnwidth}
{@{\extracolsep{\fill}}l|ccccc@{}}
\toprule
Model & Acc@GQA & mIoP & IoP@0.5 & mIoU & IoU@0.5 \\
\midrule
VIOLETv2~\cite{fu2023violetv2}
& 12.8 & 23.6 & 23.3 & 3.1 & 1.3 \\

Temp[CLIP] NG+~\cite{xiao2024can}
& 16.0 & 25.7 & 25.5 & 12.1 & 8.9 \\

SeViLA~\cite{yu2023sevila}
& 16.6 & 29.5 & 22.9 & 21.7 & 13.8 \\

LangRepo~\cite{kahatapitiya2024langrepo}
& 17.1 & 31.3 & 28.7 & 18.5 & 12.2 \\

FrozenBiLM NG+~\cite{xiao2024can}
& 17.5 & 24.2 & 23.7 & 9.6 & 6.1 \\

VideoStreaming~\cite{qian2024streaming}
& 17.8 & 32.2 & 31.0 & 19.3 & 13.3 \\

LLoVi~\cite{zhang2024llovi}
& 24.3 & \textbf{37.3} & \textbf{36.9} & 20.0 & 15.3 \\

Grounded-VideoLLM~\cite{wang2024grounded}
& 26.7 & 34.5 & 34.4 & 21.1 & 18.0 \\

\rowcolor{gray!15}
Ours
& \textbf{28.4} & 36.8 & 35.9
& \textbf{23.2} & \textbf{19.9} \\
\bottomrule
\end{tabular*}
\end{table}

\begin{table}[t]
\centering
\footnotesize
\setlength{\tabcolsep}{2.0pt}
\renewcommand{\arraystretch}{1.03}
\caption{\textbf{Open-ended VideoQA results.}
We report answer accuracy (Acc.) and GPT-based evaluation score.}
\label{tab:openended_videoqa}
\begin{tabular*}{\columnwidth}
{@{\extracolsep{\fill}}l|cccccc@{}}
\toprule
\multirow{2}{*}{Model}
& \multicolumn{2}{c}{MSVD-QA}
& \multicolumn{2}{c}{MSRVTT-QA}
& \multicolumn{2}{c}{ANet-QA} \\
& Acc. & Score & Acc. & Score & Acc. & Score \\
\midrule
Video-LLaMA~\cite{zhang2023videollama}
& 51.6 & 2.5 & 29.6 & 1.8 & 12.4 & 1.1 \\

Video-ChatGPT~\cite{maaz2024videochatgpt}
& 64.9 & 3.3 & 49.3 & 2.8 & 35.2 & 2.7 \\

VideoChat2~\cite{li2024mvbench}
& 70.0 & 3.9 & 54.1 & 3.3 & 49.1 & 3.3 \\

ST-LLM~\cite{liu2024stllm}
& 74.6 & 3.9 & 63.2 & 3.4 & 50.9 & 3.3 \\

Momentor~\cite{qian2024momentor}
& 68.9 & 3.6 & 55.6 & 3.0 & 40.8 & 3.2 \\

Grounded-VideoLLM~\cite{wang2024grounded}
& 76.3 & 4.1 & 60.3 & 3.6 & 56.8 & 3.5 \\

\rowcolor{gray!15}
Ours
& 78.0 & \textbf{4.3}
& 62.1 & \textbf{3.7}
& \textbf{58.4} & \textbf{3.6} \\
\bottomrule
\end{tabular*}
\end{table}

\begin{table}[t]
\centering
\footnotesize
\setlength{\tabcolsep}{2.8pt}
\renewcommand{\arraystretch}{1.05}
\caption{\textbf{Cumulative component ablation on Charades-STA.}}
\label{tab:ablt_component}
\resizebox{\columnwidth}{!}{%
\begin{tabular}{lccc|cccc}
\toprule
Variant & Diff. & Entity & Mix.
& R@0.3 & R@0.5 & R@0.7 & mIoU \\
\midrule
Base
& & & &
54.2 & 36.4 & 19.7 & 36.8 \\

+ Diffusion
& \checkmark & & &
56.1 & 41.3 & 20.9 & 39.5 \\

+ Entity Conditioning
& \checkmark & \checkmark & &
57.3 & 43.2 & 22.1 & 42.1 \\

+ Mixed Tokens
& \checkmark & \checkmark & \checkmark &
\textbf{58.7} & \textbf{45.2}
& \textbf{23.0} & \textbf{43.5} \\
\bottomrule
\end{tabular}%
}
\end{table}

\subsection{Experimental Setup}

\noindent\textbf{Datasets and Metrics.}
We evaluate Diffusion-Grounded VideoLLM on temporal video grounding,
grounded video question answering, and open-ended video question answering.
For temporal grounding, we use Charades-STA~\cite{gao2017tall} and
DiDeMo~\cite{hendricks2017didemo}, and report R@1 at IoU thresholds
$\{0.3,0.5,0.7\}$ together with mean Intersection over Union (mIoU).
For grounded VideoQA, we evaluate on NExT-GQA~\cite{xiao2024can} using
Acc@GQA, mIoP, IoP@0.5, mIoU, and IoU@0.5.
For open-ended VideoQA, we use MSVD-QA, MSRVTT-QA, and ActivityNet-QA,
and report answer accuracy and GPT-based evaluation score.
Unless otherwise stated, comparison results are taken from the
corresponding cited works.

\noindent\textbf{Training and Implementation.}
We initialize our model from the released Phi-3.5-3.8B
Grounded-VideoLLM checkpoint~\cite{wang2024grounded}.
Following its temporal-grounding training protocol, we train the proposed
modules on a mixture of temporal grounding and video instruction data,
including the grounding data adopted from VTimeLLM, Momentor, and
HawkEye, the Grounded-VideoQA training set, and subsets of VideoChat2
and ShareGPT4Video.
We do not use Charades-STA annotations during this adaptation.
The grounding model, video diffusion backbone, and base language model
remain frozen, while feature projections and LoRA parameters are optimized.
We train for three epochs with AdamW using a learning rate of
$10^{-4}$.
Videos are sampled at 1\,fps, retaining their source timestamps.
Unless otherwise specified, we use $N_t=8$ temporal bins, $N_o=4$ object slots, persistent entity tracking with $K=3$, and diffusion features
extracted at 10\% of the denoising trajectory, within the early 20\% range used for feature extraction.

\begin{table}[t]
\centering
\footnotesize
\setlength{\tabcolsep}{3.3pt}
\renewcommand{\arraystretch}{1.05}
\caption{\textbf{Entity-conditioning pathway ablation on Charades-STA.}}
\label{tab:entity_path}
\resizebox{\columnwidth}{!}{%
\begin{tabular}{cc|cccc}
\toprule
Object Tokens & Mask Conditioning
& R@0.3 & R@0.5 & R@0.7 & mIoU \\
\midrule
& &
56.1 & 41.3 & 20.9 & 39.5 \\

\checkmark &
& 56.6 & 41.9 & 21.3 & 40.3 \\

& \checkmark
& 57.0 & 42.6 & 21.8 & 41.5 \\

\checkmark & \checkmark
& \textbf{57.3} & \textbf{43.2}
& \textbf{22.1} & \textbf{42.1} \\
\bottomrule
\end{tabular}%
}
\end{table}

\begin{table}[t]
\centering
\footnotesize
\setlength{\tabcolsep}{5.2pt}
\renewcommand{\arraystretch}{1.05}
\caption{\textbf{Effect of diffusion feature extraction point on Charades-STA.}}
\label{tab:diffusion_state}
\begin{tabular*}{\columnwidth}
{@{\extracolsep{\fill}}c cccc@{}}
\toprule
Extraction Point & R@0.3 & R@0.5 & R@0.7 & mIoU \\
\midrule
10\%
& \textbf{58.7} & \textbf{45.2}
& \textbf{23.0} & \textbf{43.5} \\

30\%
& 57.2 & 41.9 & 22.0 & 41.4 \\

50\%
& 56.7 & 40.6 & 20.8 & 40.3 \\
\bottomrule
\end{tabular*}
\end{table}

\subsection{Main Results}

\noindent\textbf{Temporal Video Grounding.}
Table~\ref{tab:tsg_results} compares our model with representative
VideoLLM-based temporal grounding approaches.
On Charades-STA, our 3.8B model obtains 43.5 mIoU, compared with
36.8 for Grounded-VideoLLM.
R@0.5 increases from 36.4 to 45.2 and R@0.7 from 19.7 to 23.0.
Compared with the 7B LLaVA-ST, our model obtains 43.5 versus 42.4 mIoU
and 45.2 versus 44.8 R@0.5.

On DiDeMo, our model obtains 38.2 mIoU, 40.0 R@0.5, and 20.5 R@0.7,
compared with 32.0, 33.4, and 15.9 for Grounded-VideoLLM.
The gains on both benchmarks support the usefulness of the proposed
representation across temporal grounding datasets.

\noindent\textbf{Grounded VideoQA.}
Table~\ref{tab:grounded_videoqa} reports joint answer prediction and
temporal evidence localization on NExT-GQA.
Our model obtains 28.4 Acc@GQA, compared with 26.7 for
Grounded-VideoLLM.
It also increases mIoU from 21.1 to 23.2 and IoU@0.5 from 18.0 to
19.9, while obtaining 36.8 mIoP and 35.9 IoP@0.5.
The improvements in joint answer-grounding accuracy and temporal
overlap support the usefulness of the representation for grounded VideoQA.

\noindent\textbf{Open-ended VideoQA.}
Table~\ref{tab:openended_videoqa} further evaluates general
video-language understanding.
Relative to Grounded-VideoLLM, accuracy increases from 76.3 to 78.0
on MSVD-QA, from 60.3 to 62.1 on MSRVTT-QA, and from 56.8 to 58.4
on ActivityNet-QA.
The corresponding GPT-based scores are 4.3, 3.7, and 3.6.
These results show that the grounding-oriented representation retains
general VideoQA capability across the three benchmarks.

\subsection{Ablation Studies}

All ablations, including Base, use the same continued-training protocol
and are evaluated on Charades-STA. Diffusion variants use the same
representation-alignment objective; extraction-position comparisons
follow the same training protocol.

\noindent\textbf{Component Analysis.}
Table~\ref{tab:ablt_component} progressively introduces the three main
components of the framework. Adding diffusion-derived temporal features with representation alignment
improves mIoU from 36.8 to 39.5 and R@0.5 from 36.4 to 41.3.
Entity conditioning further increases mIoU to 42.1, indicating that
query-relevant object information provides additional guidance beyond
the diffusion representation alone. Finally, concatenating the visual, text, and diffusion feature sequences
through the mixed interface raises mIoU to 43.5 and R@0.5 to 45.2. The cumulative trend shows that
the three components contribute at successive stages of the grounding
pipeline.

Before introducing the final mixed interface, Table~\ref{tab:entity_path} separates the two entity-conditioning pathways. Object tokens increase mIoU from 39.5 to 40.3, while mask conditioning provides a larger gain to 41.5. Combining both pathways further improves mIoU to 42.1, showing complementary contributions from explicit entity tokens and mask-conditioned diffusion features.

\noindent\textbf{Diffusion Features.}
Earlier diffusion states perform better in our setting. Extracting features at 10\% of the denoising trajectory yields 43.5 mIoU and 45.2 R@0.5, compared with 41.4/41.9 at 30\% and 40.3/40.6 at 50\%. Among the tested positions, the early readout gives the strongest grounding performance.

\begin{table}[t]
\centering
\footnotesize
\setlength{\tabcolsep}{4.0pt}
\renewcommand{\arraystretch}{1.05}
\caption{\textbf{Effect of persistent entity tracking on Charades-STA.}}
\label{tab:tracking}
\begin{tabular*}{\columnwidth}
{@{\extracolsep{\fill}}l cccc@{}}
\toprule
Tracking Strategy & R@0.3 & R@0.5 & R@0.7 & mIoU \\
\midrule
Frame-wise grounding
& 56.1 & 40.0 & 21.1 & 39.9 \\

Tracking, $K=1$
& 57.0 & 41.8 & 21.8 & 41.2 \\

Tracking, $K=2$
& 58.0 & 43.6 & 22.5 & 42.5 \\

Tracking, $K=3$
& \textbf{58.7} & \textbf{45.2}
& \textbf{23.0} & \textbf{43.5} \\
\bottomrule
\end{tabular*}
\end{table}

\noindent\textbf{Entity Tracking.}
Table~\ref{tab:tracking} evaluates the effect of temporal persistence
during entity tracking. Frame-wise grounding obtains 39.9 mIoU,
whereas introducing persistent tracking increases mIoU to 41.2, 42.5,
and 43.5 for $K=1,2,3$, respectively. R@0.5 follows the same trend,
increasing from 40.0 to 45.2. These results support persistent tracking as an auxiliary conditioning
strategy for temporal grounding.

\begin{table}[t]
\centering
\footnotesize
\setlength{\tabcolsep}{4.0pt}
\renewcommand{\arraystretch}{1.05}
\caption{\textbf{Effect of representation budget on Charades-STA.}}
\label{tab:token_budget}
\begin{tabular*}{\columnwidth}
{@{\extracolsep{\fill}}lc cccc@{}}
\toprule
Setting & Value & R@0.3 & R@0.5 & R@0.7 & mIoU \\
\midrule
\multirow{3}{*}{$N_t$}
& 4
& 57.0 & 41.6 & 21.6 & 41.2 \\
& 8
& \textbf{58.7} & \textbf{45.2}
& \textbf{23.0} & \textbf{43.5} \\
& 16
& 58.1 & 44.1 & 22.6 & 42.8 \\
\midrule
\multirow{3}{*}{$N_o$}
& 2
& 57.5 & 42.9 & 22.0 & 42.0 \\
& 4
& \textbf{58.7} & \textbf{45.2}
& \textbf{23.0} & \textbf{43.5} \\
& 8
& 58.3 & 44.5 & 22.7 & 43.0 \\
\bottomrule
\end{tabular*}
\end{table}

\noindent\textbf{Representation Budget.}
Table~\ref{tab:token_budget} studies the numbers of temporal bins and object slots. Increasing $N_t$ from 4 to 8 improves mIoU from 41.2 to 43.5, while using 16 bins gives 42.8. Similarly, $N_o=4$ obtains 43.5 mIoU compared with 42.0 and 43.0 for 2 and 8 object slots. We therefore use $N_t=8$ and $N_o=4$ in all remaining experiments.

\section{Conclusion}
We presented \method, which conditions early video diffusion features on query-relevant tracks and combines visual, text, and diffusion features for timestamp-aware language reasoning. Experiments and ablations support the approach for temporal grounding and grounded VideoQA, including the contributions of entity conditioning and feature concatenation.

\section{Acknowledgments}
This research received no external funding. The authors declare no conflicts of interest.
\begingroup
\urlstyle{same}
\let\originalthebibliography\thebibliography
\renewcommand{\thebibliography}[1]{%
  \originalthebibliography{#1}%
  \setlength{\itemsep}{0pt}%
  \setlength{\parsep}{0pt}%
}
\bibliographystyle{IEEEbib}
\bibliography{main}
\endgroup
\end{document}